\documentclass[runningheads]{llncs}
\usepackage{graphicx,amsmath,bm,longtable,todonotes,url,microtype}
\usepackage[section]{placeins}

\def\trainn{p}

\begin{document}
	
	\title{Hierarchical forecast reconciliation with~machine~learning}
	
	\author{
		Evangelos Spiliotis \inst{1}
		Mahdi Abolghasemi\inst{2}
		Rob J Hyndman\inst{3}
		Fotios~Petropoulos\inst{4}
		Vassilios Assimakopoulos\inst{5}
	}
	
	\authorrunning{E. Spiliotis et al.}
	
	\institute{Forecasting and Strategy Unit, School of Electrical and Computer Engineering, National Technical University of Athens, Greece.
		\email{spiliotis@fsu.gr} \and
		Department of Data Science \& AI, Monash University, Melbourne VIC 3800, Australia.
		\email{mahdi.abolghasemi@monash.edu} \and
		Department of Econometrics and Business Statistics, Monash University, Melbourne VIC 3800, Australia.
		\email{rob.hyndman@monash.edu} \and
		School of Management, University of Bath, Bath, UK.
		\email{f.petropoulos@bath.ac.uk} \and
		Forecasting and Strategy Unit, School of Electrical and Computer Engineering, National Technical University of Athens, Greece.
		\email{vassim@central.ntua.gr}}
	\maketitle
	
	\begin{abstract}
		Hierarchical forecasting methods have been widely used to support aligned decision-making by providing coherent forecasts at different aggregation levels. Traditional hierarchical forecasting approaches, such as the bottom-up and top-down methods, focus on a particular aggregation level to anchor the forecasts. During the past decades, these have been replaced by a variety of linear combination approaches that exploit information from the complete hierarchy to produce more accurate forecasts. However, the performance of these combination methods depends on the particularities of the examined series and their relationships. This paper proposes a novel hierarchical forecasting approach based on machine learning that deals with these limitations in three important ways. First, the proposed method allows for a non-linear combination of the base forecasts, thus being more general than the linear approaches. Second, it structurally combines the objectives of improved post-sample empirical forecasting accuracy and coherence. Finally, due to its non-linear nature, our approach selectively combines the base forecasts in a direct and automated way without requiring that the complete information must be used for producing reconciled forecasts for each series and level. The proposed method is evaluated both in terms of accuracy and bias using two different data sets coming from the tourism and retail industries. Our results suggest that the proposed method gives superior point forecasts than existing approaches, especially when the series comprising the hierarchy are not characterized by the same patterns.
		
	\end{abstract}
	
	\section{Introduction and background} \label{literaturehierarchy}
	
	Accurate forecasting helps decision making, especially when the future is uncertain. For example, forecasting the future demand of stock keeping units (SKUs) helps in managing a supply chain, and forecasting tourist arrivals helps in capacity planning.
	
	Frequently, the time series to be forecast are naturally organized in hierarchical structures. For instance, although the demand for an SKU could be recorded on a store-by-store level, it could also be aggregated to give the demand on a regional or national level. At the same time, the demand of similar SKUs could be included in the demand of larger categories of products. These structures led to the
	development of hierarchical forecasting (HF) approaches. Such approaches are proposed in the cross-sectional \cite{athanasopoulos2009hierarchical,hyndman2011optimal,wickramasuriya2019optimal}, temporal \cite{athanasopoulos2017forecasting}, and cross-temporal domains \cite{KOURENTZES2019393,Spiliotis2020-hj}.
	
	The observed demands at each level will always add up to the observed demand at higher levels. It is usually desirable that the same holds true for forecasts --- that is, that the aggregate of the forecasts at a lower level is equal to the forecast of the aggregates at a higher level. This property is known as forecasting ``coherence'' \cite{Athanasopoulos2020-cx}. If forecasting at the different levels is done independently, we usually have forecast incoherence --- the forecasts do not add up.
	
	Until the late 2010s, the problem of forecast incoherence was bypassed by modelling and producing forecasts on a single hierarchical level:
	
	\begin{itemize}
		\item Some researchers \cite{dangerfield1992top,Zellner2000-vi} have argued for generating forecasts only on the lowest, most granular level of a hierarchy. If forecasts are needed at higher levels, these are not produced directly using the aggregated information; instead, the lower level forecasts are summed up. This approach is known as ``bottom-up'' (BU). The BU approach can be more suitable for short-term operational decisions, such as logistics and production planning \cite{kahn1998revisiting}. A downside of the BU approach is the difficulty to model each bottom level series due to the high level of noise and computational concerns in the case of large hierarchies \cite{Gross1990-bl,athanasopoulos2009hierarchical}.
		\item Other researchers \cite{Gross1990-bl,Fliedner1999} have suggested that only the top level of a hierarchy be directly forecasted, and then the forecasts are disaggregated to the lower levels using historical or forecasted \cite{athanasopoulos2009hierarchical} proportions. This approach is known as ``top-down'' (TD)\@. TD is more appropriate when strategic plans and decisions such as budgeting are made. TD generally requires fewer resources and modeling decisions, with forecasts being made on a single (top) series. However, the accuracy of the forecasts drops at lower levels of the hierarchy, due to the information loss incurred while aggregating the lower-level data to the higher aggregation levels.
		\item A solution between BU and TD is offered by the ``middle-out'' (MO) approach. In MO, forecasts are produced on an intermediate level of the hierarchy. Lower and higher level forecasts are derived by disaggregation and aggregation of the MO forecasts respectively.
	\end{itemize}
	
	The BU, MO, and TD approaches are myopic in the sense that they focus on a particular aggregation level to produce forecasts, thus ignoring some useful information \cite{pennings2017integrated} available at other levels. In the last 12 years, hierarchical forecasting approaches have significantly evolved to include combination (COM) approaches that directly tackle the challenge of coherence. COM approaches have the advantage of using the information from all hierarchical levels to produce forecasts. These forecasts are consequently combined, using weights that are obtained either statistically (see \cite{athanasopoulos2009hierarchical,hyndman2011optimal,wickramasuriya2019optimal}) or empirically (see the cross-validation approach in \cite{Jeon2019-xo}). Simpler combinations based on equal weights have also been shown to be useful under some settings \cite{Abouarghoub2018-wz}. The application of hierarchical combination approaches has one direct advantage: it renders forecasts across hierarchical levels coherent, a property that is desirable in aligning decision making across the different functions of an organization. Apart from its direct benefits, more often than not, COM also results in superior forecasting performance compared to simpler HF approaches.
	
	The hierarchical combination approaches that have been explored so far in the literature are linear in nature. The only existing non-linear approach in HF, proposed by \cite{abolghasemi2019machine}, uses ML models under the MO approach to dynamically forecast the proportions of the child nodes from their parent. However, this approach exploits information only from the parent node, ignoring the rest of the nodes that could be useful for obtaining more accurate results.
	
	In all four aforementioned approaches (BU, TD, MO, and COM), the base forecasts can be generated using any statistical or judgmental forecasting method. Indeed, the method of choice might differ depending on the aggregation level of focus and the data availability. Popular choices include univariate forecasting models, such as exponential smoothing (ETS) and AutoRegressive Integrated Moving Average (ARIMA) models. However the baseline models could also allow for exogenous information, which may be crucial in, for example, retail settings where promotions occur often. Moreover, \cite{Spiliotis2019-hg} showed that combining the forecasts across methods in order to obtain more accurate base forecasts will also increase the performance of the final, reconciled hierarchical forecasts. The efficacy of the different HF approaches depends on the time series features, the level of forecasting, the forecasting horizon, the structure of the hierarchy, and the relationships of the series. We may consider these variables when choosing the most appropriate HF approach \cite{Fliedner1999,Fliedner2001,gross1990disaggregation,nenova2016determining,abolghasemi2019machine}.
	
	In this paper we offer a non-linear perspective to the problem of hierarchical reconciliation and forecast coherence. We propose the use of Machine Learning (ML) techniques to derive the combination weights for the forecasts across the various aggregation levels. We focus on two ML algorithms that have been shown to perform well in time series contexts and cross-learning: Random Forests (RF) and XGBoost (XGB). Such decision tree algorithms allow the exploitation of non-linear relationships across a number of series. This is particularly useful in hierarchical structures, especially when exogenous variables are available only on some of the hierarchical levels, the series are not all characterized by the same patterns, or the relationships of the series change through time. The contributions of this paper are threefold:
	
	\begin{itemize}
		\item We propose non-linear approaches to the problem of hierarchical forecast reconciliation. These approaches are more general compared to their linear counterparts and are expected to enhance the forecasting performance across all hierarchical levels.
		\item The majority of the existing HF reconciliation approaches are, strictly speaking, designed to result in coherence under particular assumptions, with improvements in terms of forecasting performance being an all-welcome side-effect. In contrast, our proposed approaches structurally combine the objectives of post-sample empirical forecast accuracy and coherence in the training phase of the ML algorithm. The only other approach in the literature that has this property is HF via cross-validation \cite{Jeon2019-xo}. Other methods that optimize forecast accuracy under the constraint of linear coherence, like the one proposed by \cite{wickramasuriya2019optimal}, do so using the one-step-ahead in-sample errors of the baseline forecasting methods, which may not be representative of post-sample accuracy.
		\item Unlike existing HF approaches, our proposed approaches selectively combine the forecasts across the different nodes of the hierarchy in a direct and automated way, without requiring that all forecasts need to be used.
	\end{itemize}
	
	We benchmark the performance of the proposed ML HF approach against various state-of-the-art methods on two datasets coming from the tourism and retail industries, using ARIMA-like approaches to estimate the base forecasts. Our results suggest that ML reconciliation approaches are superior to existing, linear ones, both in terms of accuracy and bias.
	
	The remainder of the paper is organized as follows. Section~\ref{HFmodels} describes the most popular HF methods found in the literature, while Section~\ref{dynamicproposalhier} presents the proposed ML reconciliation approach. Section~\ref{hierarchicalresults} presents the two datasets used for the empirical evaluation of the proposed method, describes the experimental set-up, and discusses our results and findings. Section~\ref{hierarchicalconclusion} concludes the paper.
	
	\section{Hierarchical forecasting models} \label{HFmodels}
	
	In this section, we discuss the TD, BU, and COM methods as three well-established HF approaches. The following indices, notations, and parameters are used throughout this paper:
	
	\begin{longtable}{r@{ = }p{10.6cm}}
		$m$                     & total number of series in the hierarchy; \\
		$m_i$                   & total number of the series for level $i$; \\
		$k$                     & total number of the levels in hierarchy; \\
		$n$                     & number of the observations in each series; \\
		$Y_{x,t}$               & the $t^{th}$ observation of series $Y_x$; \\
		$\hat{Y}_{x,n} (h)$     & $h$-step-ahead independent base forecast of series $Y_x$ based on $n$ observations; \\
		$\bm{Y}_{i,t}$          & the vector of all observations at level $i$; \\
		$\hat{\bm{Y}}_{i,t}(h)$ & $h$-step-ahead forecast at level $i$; \\
		$\bm{Y}_t$              & a column vector including all observations; \\
		$\hat{\bm{Y}}_n (h)$    & $h$-step-ahead independent base forecast of all series based on $n$ observations; \\
		$\tilde{\bm{Y}}_n (h)$  & the final reconciled forecasts of all series. \\
	\end{longtable}
	
	\addtocounter{table}{-1}
	
\begin{figure}
		\centerline{\includegraphics[scale=0.55]{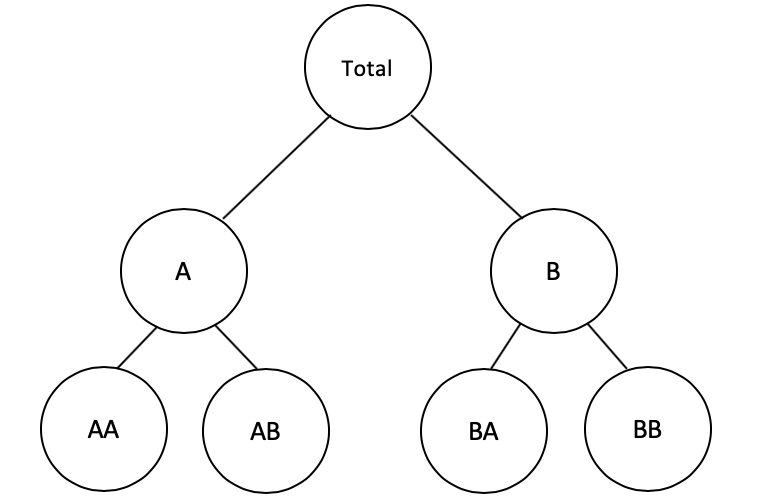}}
		\caption{A three level hierarchical structure}
		\label{htsstructure}
	\end{figure}

	The hierarchical time series can be expressed as $\bm{Y}_t = \bm{S} \bm{Y}_{k,t}$, where $\bm{S}$ is a summing matrix of order $m \times m_k$ that aggregates the bottom level series. Consider the hierarchy shown in Figure~\ref{htsstructure} that shows a three level hierarchy.
	
	The hierarchy shown in Figure~\ref{htsstructure} can be expressed as:
	\begin{equation*}
	\left[
	\begin{array}{c}
	Y_t \\
	Y_{A,t} \\
	Y_{B,t}\\
	Y_{AA,t} \\
	Y_{AB,t}\\
	Y_{BA,t} \\
	Y_{BB,t} \\
	\end{array}
	\right] =
	\left[
	\begin{array}{@{}*{4}{c}@{}}
	1 & 1 & 1        & 1 \\
	1 & 1 & 0        & 0 \\
	0 & 0 & 1        & 1 \\
	&   & \bm{I}_4 & \\
	\end{array}
	\right]
	\times
	\left[
	\begin{array}{c}
	Y_{AA,t} \\
	Y_{AB,t} \\
	Y_{BA,t} \\
	Y_{BB,t} \\
	\end{array}
	\right]
	\end{equation*}
	The various HF approaches can then be expressed with a unified structure $\tilde{\bm{Y}}_n (h)= \bm{S} \bm{G} \hat{\bm{Y}}_n(h)$, where $\bm{G}$ is a matrix of order $m \times m_k$ which elements depend on the type of the HF method used \cite{fpp2}.
	
	\subsection{Bottom-up}\label{buhier}
	
	The BU approach considers just the base forecasts produced on the bottom level of the hierarchy and sums them appropriately to obtain forecasts at higher levels. In this approach, $\bm{G}= [\bm{0}_{m_k \times (m - m_k)}| \bm{I}_{m_k}]'$, where $\bm{0}_{i \times j}$ is an $i \times j $ null matrix. Thus, $\bm{G}$ extracts the bottom level forecasts and combines them with the summing matrix $\bm{S}$ to generate the final forecasts of the hierarchy.
	
	\subsection{Top-down}\label{tdhier}
	
	In the TD approach, base forecasts are produced just at the top level of the hierarchy and are then disaggregated to the lower levels with an appropriate factor. Gross and Sohl \cite{gross1990disaggregation} investigated 21 different disaggregation methods for the TD approach. They concluded that Equations~\eqref{td1} and~\eqref{td2} indicate two disaggregation methods that give reasonable forecasts at the bottom level.
	\begin{align}
	\label{td1}
	p_j = \frac{1}{n}\sum_{t=1}^{n} \frac{Y_{j,t}}{Y_t} \qquad j=1, \dots, m_k \\
	\label{td2}
	p_j = \frac{\sum_{t=1}^{n} Y_{j,t}}{\sum_{t=1}^{n} Y_t} \qquad j=1, \dots, m_k
	\end{align}
	In Equation~\eqref{td1}, each proportion $p_j$ reflects the average of the historical proportions of the bottom level series ${Y_{j,t}}$, while in Equation~\eqref{td2}, each proportion $p_j$ reflects the average of the historical value of the bottom level series ${Y_{j,t}}$ relative to the average value of the total aggregate ${Y_t}$. These proportions can be used to form the vector $\bm{g} =[p_1, p_2, p_3, \dots, p_{m_k}]$ so that $\bm{G}= [\bm{g} \mid \bm{0}_{m_k \times (m - 1)}]'$. In this regard, $\bm{G}$ disaggregates the forecast at the top level to the lower levels.
	
	Athanasopoulos et al.\ \cite{athanasopoulos2009hierarchical} proposed the TD forecasted proportions (TDFP) approach that disaggregates the top level forecasts based on the forecasted proportions of lower level series rather than the historical proportions. According to this method,
	$$
	p_j = \prod_{i=0}^{k-1} \frac{\hat{Y}^{(i)}_{j,n} (h)} {\sum (\hat{Y}^{(i+1)}_{j,n} (h))},
	$$
	for $j=1, \dots, m_k$, where $\hat{Y}^{(i)}_{j,n} (h)$ is the $h$-step ahead forecast of the series that corresponds to the node which is $i$ levels above $j$, and $\sum\hat{Y}_{i,n} (h)$ is the sum of the $h$-step ahead forecasts below node $i$ that corresponds directly to the node $i$. These will form the vector $\bm{g} =[p_1, p_2, p_3, \dots, p_{m_k}]$ so that $\bm{G}= [\bm{g} \mid \bm{0}_{m_k \times (m - 1)}]'$. Similarly to the rest of the TD methods, TDFP approach will generate biased forecasts even if the base forecasts are unbiased \cite{athanasopoulos2009hierarchical}.
	
	We use the \texttt{td} function in \textit{hts} package to implement the TD method \cite{htspackage} that utilizes the proportions of Equation~\eqref{td1}.
	
	\subsection{Linear combination}\label{optimalhier}
	
	The COM method uses a completely different approach for HF\@. This approach was developed over a series of papers by Hyndman et al. \cite{hyndman2011optimal,hyndman2016fast} and Wickramasuriya et al \cite{wickramasuriya2019optimal}. Let the $h$-step reconciled forecasts be given by
	$$
	\tilde{\bm{Y}}_n(h) = \bm{S} \bm{G} \hat{\bm{Y}_n}(h).
	$$
	They showed that the covariance matrix of the $h$-step-ahead reconciled forecast errors is given by
	$$
	\bm{V}_h=\text{Var}[\bm{y}_{n+h} - \tilde{\bm{Y}}_n (h)]= \bm{S}\bm{G}\bm{W}_h\bm{G}'\bm{S}',
	$$
	where $\bm{W}_h$ is the variance-covariance matrix of the $h$-step ahead base forecast errors. Moreover, they demonstrate that if the base forecasts are unbiased, these reconciled forecasts will also be unbiased if and only if $\bm{S}\bm{G}\bm{S} = \bm{S}$. Finally, they showed that the $\bm{G}$ matrix that minimizes the trace of $\bm{V}_h$ such that $\bm{S}\bm{G}\bm{S}=\bm{S}$ is given by
	$$
	\bm{G}= (\bm{S}' \bm{W}^{\dagger}_h \bm{S})^{-1} \bm{S}' \bm{W}_h^{\dagger},
	$$
	where $\bm{W}^\dagger_h$ is the generalized inverse of $\bm{W}_h$. Hence, the optimal unbiased forecasts from a linear reconciliation are given by
	$$
	\tilde{\bm{Y}}_n(h) = \bm{S} (\bm{S}' \bm{W}^{\dagger}_h \bm{S})^{-1} \bm{S}' \bm{W}_h^{\dagger} \hat{\bm{Y}_n}(h).
	$$
	This is known as MinT (minimum trace) reconciliation . Note that it can be considered a generalized least squares estimator for a corresponding regression problem.
	
	The challenge is the estimation of $\bm{W}_h$, especially for very large hierarchies, and different approximate estimates have been proposed.
	
	\begin{enumerate}
		\item Set $\bm{W}_h=k_h \bm{I}$. This is known as the OLS estimator \cite{hyndman2011optimal}. This ignores the scale of each series and the relationships between the series.
		\item Set $\bm{W}_h=k_h \text{diag}(\hat{\bm{W}}_1)$ where $\hat{\bm{W}}_1= \frac{1}{T}\sum_{t=1}^{T} \hat{\bm{e}}_T (1) \hat{\bm{e}}'_T (1)$ is the sample covariance matrix of the one-step ahead base forecast errors given by $\hat{\bm{e}}_T (1)= \bm{y}_{T+1} - \bm{\hat{y}}_T(1)$. This is known as the WLS estimator \cite{hyndman2016fast}. It ignores the relationships between the series, but takes account of the scale of each series.
		
		\item Set $\bm{W}_h=k_h \text{diag}(\bm{S1})$ where $\bm{1}$ is a unit $n$ vector. This method ignores the relationships between the series and assumes that the bottom level series have errors with equal variances \cite{athanasopoulos2017forecasting}. Because this method only depends on the structure of the hierarchy, it is known as structural scaling. It is particularly useful when residuals are not available.
		
		
		\item Set $\bm{W}_h=k_h\left(\lambda_D \hat{\bm{W}}_{1,D} + (1-\lambda_D)\hat{\bm{W}}_1\right)$. This is a shrinkage estimator with  diagonal target $\hat{\bm{W}}_{1,D} = \text{diag}(\hat{\bm{W}}_1)$, and shrinkage parameter
		$$
		\lambda_D = \frac{\sum_{i \neq j} \text{Var}(\hat{r}_{ij})}{\sum_{i \neq j} \hat{r}^2_{ij}},
		$$
		where $\hat{r}_{ij}$ is the $(i,j)^{th}$ element of the one step ahead in-sample correlation matrix \cite{schafer2005shrinkage}. The main advantage of this method is that it considers the relationships between the series.
	\end{enumerate}
	
	In this paper, we consider the latter two methods: structural scaling (COM-SS) and shrinkage (COM-SHR). We use the \texttt{MinT} function in \textit{hts} package in R to implement the COM-SHR and COM-SS methods \cite{htspackage}.
	
	\section{ML hierarchical forecasting approach} \label{dynamicproposalhier}
	
	In this section we present a ML reconciliation approach that exploits the potential of decision tree-based algorithms. It is designed to  deal with the limitations of the existing HF methods, highlighted in Section ~\ref{literaturehierarchy}, and allow for the base forecasts produced for the complete hierarchy to be effectively combined in a non-linear fashion to yield coherent forecasts. We consider the Random Forest (RF) and the XGBoost (XGB) algorithms as they are intuitively easy to understand and have shown promising results in time series forecasting, especially in applications where information is extracted from large time series data sets in order for the relationships of the series to be learned and the overall forecasting accuracy to be enhanced \cite{MONTEROMANSO202086}.
	
	\subsection{Proposed algorithm}
	
	The proposed ML reconciliation algorithm uses time series cross-validation \cite{fpp2} to measure the out-of-sample forecast accuracy, which is then used in an optimization procedure to tune the ML method.
	
	Assume a time series hierarchy that consists of $k$ levels and $m$ series, with each series of length $n$. We summarize our approach as follows.
	
	\begin{enumerate}
		\item The series are split into a series of training sets and test sets, with each training set comprising the first $\trainn<n$ observations (for $p=q,q+1,\dots,n-1$) and the corresponding test set comprising only the observations at time $\trainn+1$.
		
		\item A forecasting model is fitted to each series in each training set and one-step-ahead forecasts are produced for each test set.
		
		\item A separate ML model (either a RF or XGB) is built for predicting each of the $m_k$ bottom series of the hierarchy. The training set of each model consists of $n - \trainn$ observations and $m+1$ variables. The first $m$ variables (used as predictors or inputs) are the one-step ahead forecasts produced during the rolling origin process for the $m$ series of the hierarchy, and the last  variable (the response or target) is the actual value of the bottom-level series at the corresponding times. The loss function of the models is the sum of squared errors, and the hyper-parameters of the ML models are determined either arbitrarily by the user or through an optimization procedure.
		
		\item The complete sample of the series (all $n$ observations) is used to produce $h$-step-ahead base forecasts for the $m$ series of the hierarchy, where $h$ is the forecasting horizon of interest.
		
		\item The $m_k$ models that were built in Step 3 are used to provide forecasts for the series of the bottom level of the hierarchy, using the base forecasts produced in Step 4 as input. This process is repeated $h$ times, each time for a different forecasting horizon.
		
		\item The forecasts produced by the ML models in step 5 are aggregated (summed) so that reconciled forecasts are produced for the rest of the hierarchical levels.
	\end{enumerate}
	
\begin{figure}
		\centering
		\caption{Demonstration of the proposed Machine Learning hierarchical forecasting approach for the case of a two-level hierarchy consisting of one parent and two child nodes.}
		\includegraphics[width=\textwidth]{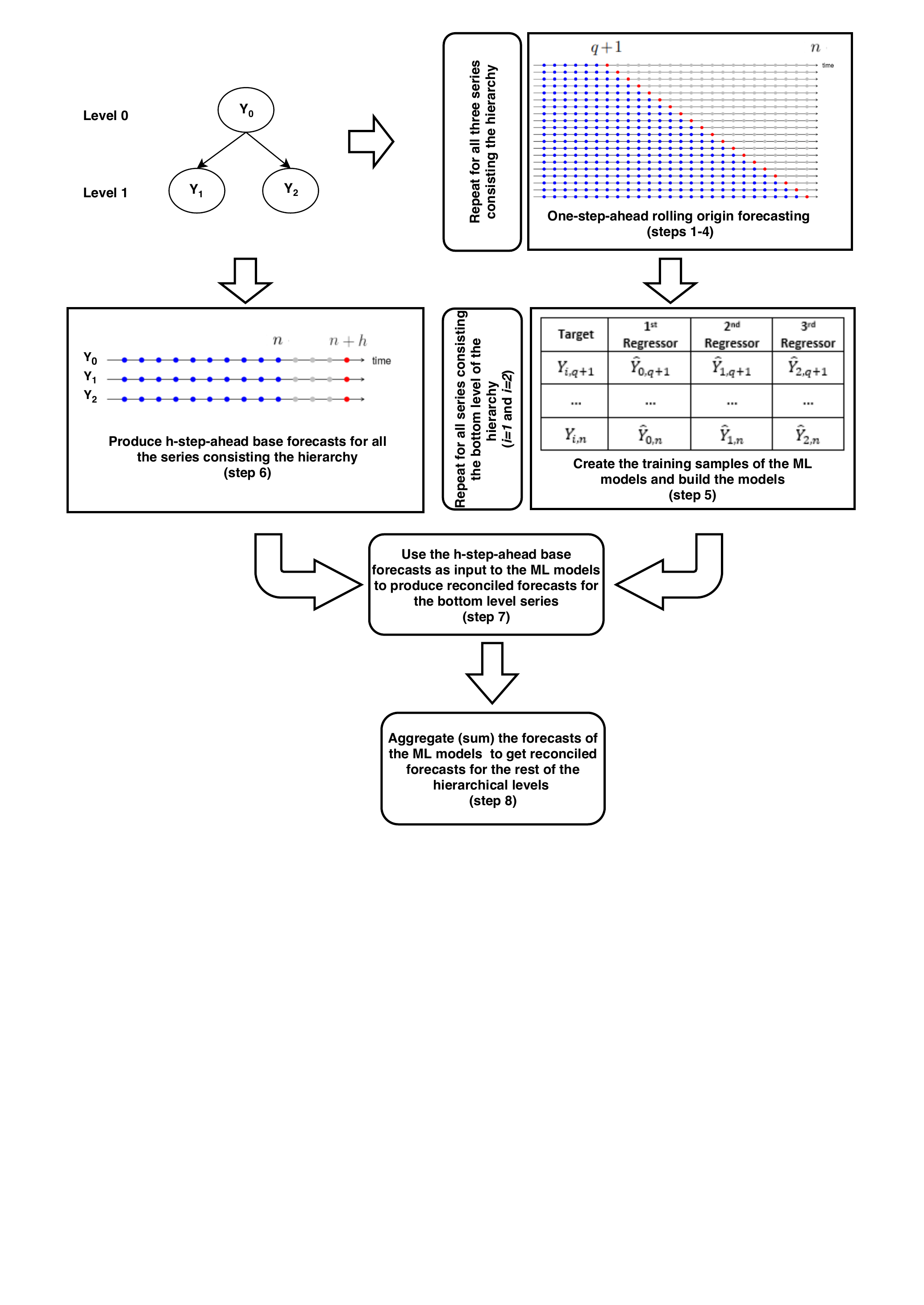}
		\label{fig:meth_approach}
	\end{figure}

	The proposed approach is demonstrated in Figure~\ref{fig:meth_approach} for the case of a simple, two-level hierarchy with one parent and two child nodes.
	
	As seen, the proposed ML HF approach provides coherent forecasts by exploiting the information available at all hierarchical levels, following the approach used by the COM methods. The main difference between the COM methods and the proposed framework is that the base forecasts are not all necessarily used for deriving the reconciled ones, being selectively handled by the ML models built for this purpose. Moreover, even if all base forecasts are to be used by the ML models, the combination of the base forecasts will be done in a non-linear fashion with the weights not being directly related to the structure of the hierarchy or the residuals reported for the individual series/levels. Most importantly, since the ML models are trained with the explicit objective of minimizing the forecasting error for each series of the bottom level of the hierarchy, the reconciliation performed may lead to more accurate forecasts when compared to standard HF methods. Finally, note that each bottom-level series is predicted by a separate ML model, meaning that the reconciliation performed is highly specialized and, therefore, able to adapt to different patterns in each series.
	
	Observe that the proposed approach is easy to generalize and is model independent. For example, a Neural Network (NN) or a Support Vector Machine (SVM) could be used to replace RF and XGB\@. Similarly, any model of choice could be used for producing the base forecasts being reconciled. Moreover, the one-step-ahead forecasts produced for constructing the training sets of the ML models, could be easily expanded to $h$-step-ahead ones to better simulate the forecasting task under examination. Our proposal of using one-step-ahead forecasts is mainly based on the fact that by increasing the forecasting horizon of the base models, the observations of the training set, i.e., $n - \trainn$, are accordingly decreased. Thus, when dealing with low frequency data (e.g., monthly or quarterly) or relatively short time series, such an approach could significantly reduce the potential of the developed ML models.
	
	The following subsections present the ML models used in this study for reconciliation. This includes information about the way the models were trained, optimized, and implemented.
	
	\subsection{XGBoost}\label{xgboostmodel}
	
	XGB is an ensembling method based on decision trees that uses a gradient boosting approach to generate unbiased and robust forecasts \cite{chen2016xgboost}. This algorithm has been applied to various forecasting and classification problems with promising results \cite{nielsen2016tree,chatzis2018forecasting,demolli2019wind}.
	
	XGB uses a number of hyper-parameters that play a critical role in generating the final forecasts. There are various techniques for optimizing these hyper-parameters, including grid search, sequential model based algorithm configuration, and Bayesian optimization. Since grid search is computationally expensive, we tuned the hyper-parameters using a Bayesian optimization approach with 10-fold cross-validation \cite{snoek2012practical}. The Bayesian approach starts with a~priori values for the hyper-parameters and then iteratively updates to identify the best values for the investigated problem. We considered intervals with different lower and upper bounds for each hyper-parameter. We set the prior values of the learning rate, \texttt{eta}, between (0.01, 0.05), \texttt{sub sample} size prior values between (0.3, 1), \texttt{colsample-bytree} prior values between (0.3, 1), \texttt{min-child weight} between (0, 10), \texttt{max-depth} between (2, 10), and \texttt{gamma} between (0, 5). The values for the maximum number of boosting iterations rolled over the range of 50 and 200. We used a linear regression model as the objective function and chose the best results by minimizing the root mean squared error (RMSE). We tuned the hyper-parameters using the \textit{rBAyesianOptimization} package for R \cite{rBayesianopt}. Due to the notable variations present in the ``Sales'' dataset (see subsection~\ref{Methodologyhier}), the optimization of the hyper-parameters was performed for each hierarchical time series and set of child-parents separately, while for the ``Tourism'' dataset we optimized the hyper-parameters for the hierarchy as a whole in order to accelerate the whole process.
	
	\subsection{Random Forest} \label{rfmodel}
	
	RF is an ensembling method that combines a large number of decision trees and takes an average of the trees to generate the final forecast \cite{breiman2002manual}. Each tree of the RF is based on a random draw from the training dataset. The trees are built using the bootstrapping method and splitting criteria in nodes. We consider the weighted variance as the splitting criteria which minimizes the sum of squared errors. This method has been successfully applied to numerous forecasting problems such as energy \cite{smyl2019machine,cheng2012random} and sales \cite{jimenez2017multi} forecasting.
	
	RF is fast to run and it only has a few hyper-parameters: the number of trees (\texttt{ntree}), node size (\texttt{nodesize}), and number of variables sampled at each split (\texttt{mtry}). Of these, the number of constructed trees is the most important feature to be tuned. The problem of optimally selecting the number of trees has been intensively discussed in the literature \cite{probst2019hyperparameters,breiman2002manual,barman2014prediction,breiman1984classification}. The main problem is that although creating more trees is computationally more demanding, it does not guarantee a better forecast. This is because each tree is trained individually and so by adding more trees, over-fitting may occur \cite{breiman2002manual}. On the other hand, since the individual trees constructed do not have the learning capacity of XGB, RF is typically more robust to outliers and over-fitting, especially for limited samples of data \cite{FRIEDMAN2002367}. The hyper-parameter \texttt{mtry} denotes the number of variables sampled at each split and controls the randomness of the model. The \texttt{nodesize} hyper-parameter determines the minimal number of observations in a terminal node to be split.
	
	We used grid search, an automated method that explores a set of different hyper-parameters values and computes the error on the validation set, with 10-fold cross-validation to find the optimal number of trees by minimizing the RMSE. We ran \texttt{ntree} on a sequence of intervals of width 5 ranging from 50 to 150 and fitted the best model using the \textit{randomForest} package for R \cite{randomForest}. We tuned the other two hyper-parameters, \texttt{mtry} and \texttt{nodesize}, using the \textit{mlr} package in R \cite{mlrpackage}. The lower and upper bound values for \texttt{mtry} were set between 2 and 6, respectively. The lower and upper bound values for \texttt{nodesize} were set on 10 and 50, respectively. Once again, the optimization of the hyper-parameters for the case of the ``Sales'' dataset was performed for each hierarchical time series and set of child-parents separately, while for the ``Tourism'' dataset it was done for the hierarchy as a whole.
	
	\section{Empirical results}\label{hierarchicalresults}
	
	\subsection{Data}\label{Methodologyhier}
	
	In order to empirically evaluate the performance of the proposed ML HF methods, we consider two different datasets, to be named the ``Tourism'' and the ``Sales'' dataset.

	\begin{table}[!htbp]
		\centering
		\caption{Number of time series per level of hierarchy in the ``Tourism'' dataset.}
		\begin{tabular}{c c}
			\textbf{Hierarchical level } & \textbf{Number of series} \\
			\hline
			Level 0                      & 1 \\
			Level 1                      & 7 \\
			Level 2                      & 27 \\
			Level 3                      & 76 \\
			\hline
			Total                        & 111
		\end{tabular}\label{table:structure_t}
	\end{table}
	
	\begin{figure}[!htbp]
		\centering
		\caption{Domestic visitor nights, measured in millions, for selected geographic divisions of Australia. A sample of indicative series is used for representing each level of the ``Tourism'' dataset.}
		\includegraphics[width=\textwidth]{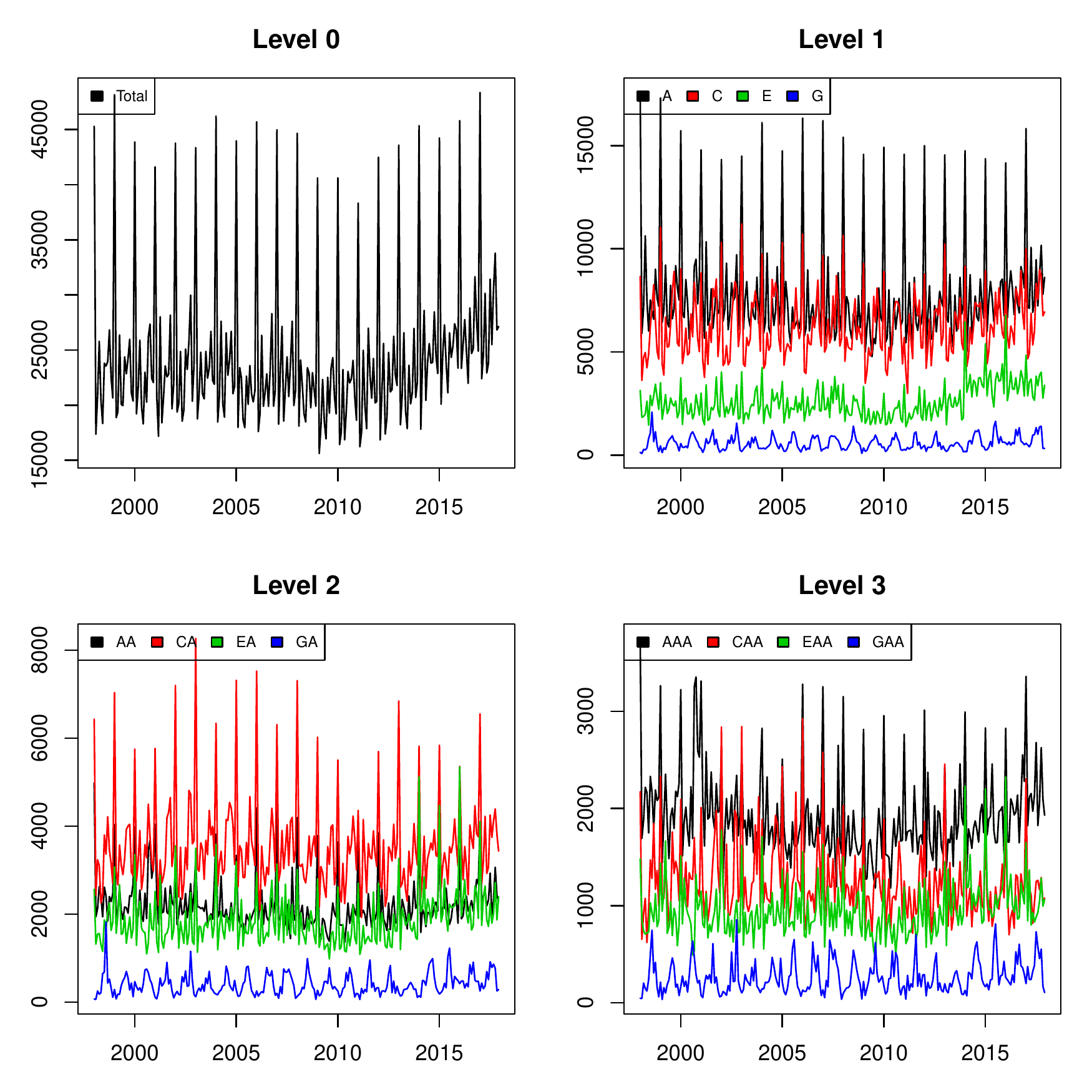}
		\label{fig:tourism_dataset_fig}
	\end{figure}
	
	The Tourism dataset involves a four-level hierarchy with the domestic visitor nights of Australia, measured in millions, across 76 regions (level 3). The regions can be grouped into 27 zones (level 2), which can be further aggregated into 7 states and territories (level 1), as well as into the total domestic visitor nights (level 0). Thus, based on these geographic divisions, the Tourism dataset comprises 111 time series. The series have a duration of 240 months (20 years) and span from January 1998 to December 2017.
	
	Table~\ref{table:structure_t} summarizes the number of series present per hierarchical level, while Figure~\ref{fig:tourism_dataset_fig} visualizes some indicative series from each level. Observe that the trend and seasonal patterns differ among the series, especially for different states and territories. Moreover, the trend of some series (e.g., A, C, and E) changes through the years, in contrast to others (e.g. G) that remain quite constant. This indicates that considering a dynamic, non-linear HF method instead of a linear one, could prove beneficial for predicting these series. For more information about the dataset, please see \cite{KOURENTZES2019393}.

	The Sales dataset involves 55 three-level hierarchies that present the sales of the cereal and breakfast products sold by a company in various locations of Australia, along with the corresponding prices. Each hierarchy refers to a different product, with the first level (level 0) representing the total sales of the manufacturer, the second level (level 1) the way these sales are disaggregated into two retailers, and the third level (level 2) the sales reported for each of the six distribution centers (DCs) used by each retailer. Thus, the Sales dataset includes 55 hierarchies, each consisting of 15 time series. The series have a duration of 120 weeks and span from September 2016 to December 2018.
	
\begin{figure}
		\centering
		\caption{Sales of an indicative cereal/breakfast product sold in Australia. The sales are presented in total, as well as per retailer and distribution center. This is an indicative example of the hierarchies involved in the ``Sales'' dataset.}
		\includegraphics[scale=0.5]{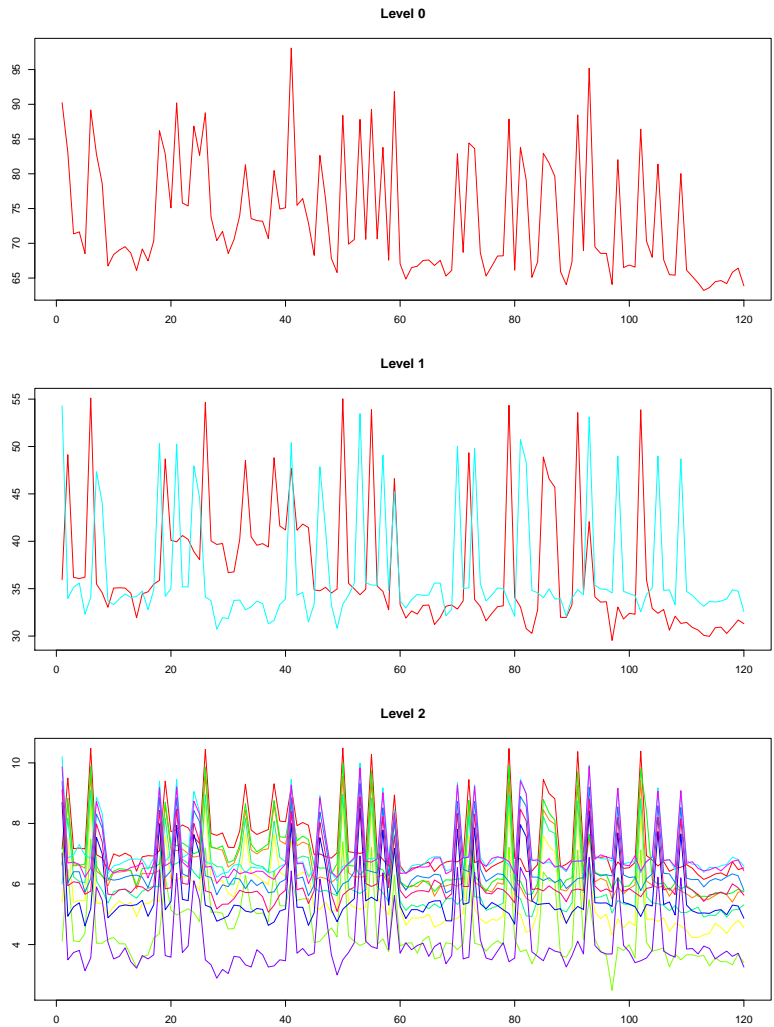}
		\label{fig:sales_dataset_fig}
	\end{figure}

	Table~\ref{table:structure_s} summarizes the number of series present per hierarchical level, while Figure~\ref{fig:sales_dataset_fig} visualizes the series of each level for one indicative product of the dataset. Note that, although the retailers display different demand patterns, DCs have a similar pattern to their retailers in terms of promotions. Moreover, different entities of the hierarchy may experience different levels of uplifts in sales. Thus, a ML HF method, which effectively captures sales variations, could be more effective for reconciling the base forecasts of these series than a traditional, linear one.
	
	\begin{table}
		\centering
		\caption{Number of time series per level of hierarchy in the ``Sales'' dataset. The hierarchical structure is the same for all 55 products of the dataset.}
		\begin{tabular}{c c}
			\textbf{Hierarchical level } & \textbf{Number of series} \\
			\hline
			Level 0                      & 1 \\
			Level 1                      & 2 \\
			Level 2                      & 12 \\
			\hline
			Total                        & 15
		\end{tabular}\label{table:structure_s}
	\end{table}
	
	\FloatBarrier
	
	\subsection{Experimental set-up}\label{sec:setup}
	
	Given that each dataset displays its own particular characteristics, we consider a different forecasting model in each case for producing the base forecasts. More specifically, for the Tourism dataset we consider ARIMA models, as implemented in the \textit{forecast} package for R \cite{hyndmanforpack}, while for the Sales dataset we use regression models with ARIMA errors (RegARIMA) using price as an regressor variable. In this respect, the effect of the promotions, which typically increase sales and drive major changes in the underlying demand behaviour \cite{Nikolopoulos2015}, is effectively taken into account. We should note that ETS \cite{Hyndman2002b} and Theta \cite{assimakopoulos2000theta} were also tested for the case of the Tourism dataset, providing similar insights to the ones reported for ARIMA\@. Thus, for reasons of brevity, and in order for the baseline models used in both cases to be similar in nature, we proceed by reporting the results only for the ARIMA models.
	
	We use the ARIMA models for producing 12-step-ahead forecasts for the monthly series of the Tourism dataset and the RegARIMA models for producing 8-step-ahead forecasts for the weekly series of the Sales dataset. Then, we utilize the BU, TD (using the disaggregation method depicted in Equation~\ref{td1}), COM-SS, and COM-SHR methods for reconciling the base forecasts of these models, as well as the ML-RF and ML-XGB HF methods proposed in this study. The first four methods are used as benchmarks as they involve both standard and state-of-the-art HF approaches.
	
	We evaluate the forecasting performance of the HF methods both in terms of accuracy (absolute deviation of the forecasts around the true values) and bias (consistent distance observed between the forecasts and the true values), using the mean absolute scaled error (MASE) \cite{Hyndman2006a}, as well as the root mean squared scaled error (RMSSE) and absolute mean scaled error (AMSE). The measures can be calculated as
	\begin{align*}
	\text{MASE}  & = \frac{n-s}{h} \frac{ \sum\nolimits_{t=n+1}^{n+h} {|y_{t}-f_{t}|} } {\sum\nolimits_{t=s+1}^{n} |y_{t}-y_{t-s}|}, \\
	\text{RMSSE} & = \frac{n-s}{h} \sqrt { \frac{ \sum\nolimits_{t=n+1}^{n+h} {(y_{t}-f_{t})^2} } {\sum\nolimits_{t=s+1}^{n} (y_{t}-y_{t-s})^2} }, \\
	\text{AMSE}  & = \frac{n-s}{h} \frac{ {|\sum\nolimits_{t=n+1}^{n+h} {(y_{t}-f_{t})} }| } {\sum\nolimits_{t=s+1}^{n} |y_{t}-y_{t-s}|},
	\end{align*}
	where $y_t$ and $f_t$ are the observation and the forecast for period $t$, $n$ is the sample size (observations used for training the forecasting model), $s$ is the length of the seasonal period, and $h$ is the forecasting horizon. In all cases, lower values indicate better forecasts.
	
	Note that all measures are scale-independent, meaning that averaging across series is possible. Moreover, given that the median minimizes the sum of the absolute errors \cite{Neil199044138}, while the mean minimizes the sum of the squares of these errors \cite{KOLASSA2016788}, MASE and RMSSE are appropriate for evaluating the accuracy of the examined HF method in approximating the median and the mean of the future values, respectively. Accordingly, AMSE is appropriate for measuring the bias of the reconciled forecasts.
	
	In order for our results to represent reality as close as possible and approximate the actual performance of the examined HF methods in a long-term run, we consider the rolling-origin evaluation approach \cite{TASHMAN2000437}. According to this approach, the first $N$ observations of each series are used for producing $h$-step-ahead forecasts, with the following $N+1 \dots N+h$ observations used for evaluating them. Then, the forecasting origin is increased by one and new forecasts are produced from the updated origin, this time using $N+1$ observations for training the forecasting model and the following $N+2 \dots N+h+1$ ones for testing. This process is repeated $K$ times, until there are no observations left for evaluating the forecasts, i.e., while $N+h+K-1 \leq n$.
	
	Given that the length and the frequency of the series of the two datasets differ, we consider a different, yet indicative implementation of the rolling-origin evaluation approach per case. Specifically, in the Tourism dataset we begin to produce forecasts at the end of the $14^{th}$ year of the dataset ($N_1=168$ months) and use the remaining 6 years for testing, thus performing a total of $K_1=61$ evaluations. Accordingly, in the Sales dataset, we start producing forecasts at the end of the $1^{st}$ year of the dataset ($N_2=52$ weeks) and use the remaining 60 weeks of each sales time series for testing, thus performing a total of $K_2=61 \times55=3355$ evaluations. The overall performance of the HF methods in each dataset is computed by averaging the scores reported across all $K_1$ and $K_2$ evaluation periods.
	
	Note that in order for the ML HF methods to be effectively trained to derive accurate reconciled forecasts when provided with a set of base forecasts, we require a dataset that includes an adequate sample of past, actual time series values (target variables) and the corresponding base forecasts produced for these periods by the forecasting model (regressor variables). In order to obtain such a dataset, we produce multiple one-step-ahead forecasts in a rolling-origin fashion, starting from an initial point, $\trainn$, and finishing at the forecast origin considered in each repetition of the rolling-origin evaluation approach, as described in Section~\ref{dynamicproposalhier} (steps 1--4). We set $\trainn$ equal to $\trainn_1=60$ and $\trainn_2=26$ for the Tourism and Sales dataset, respectively, so that a reasonable amount of full seasonal periods is available for producing the base forecasts to be used for training the ML HF methods. In this regard, in the first evaluation performed for the Tourism dataset, a sample of $N_1-\trainn_1=108$ records will be available for training the ML HF methods, with the records becoming $N_1+61-1-\trainn_1=168$ in the last evaluation. Accordingly, a sample of $N_2-\trainn_2=26$ records will be available in the first evaluation of the Sales dataset for each of the 55 hierarchies, with their length reaching $N_2+60-1-\trainn_2=85$ records in the last evaluation.
	
	\subsection{Results}
	
	Tables~\ref{table:results_t} and~\ref{table:results_s} summarize the performance of the HF methods considered in this study in terms of accuracy (MASE and RMSSE) and bias (AMSE) for the Tourism and the Sales dataset, respectively. The first column of each table indicates the HF methods considered, while the rest of the columns present the performance of the method for each aggregation level separately, as well as across all levels (average of measure values reported for Level 0 to Level $k$). All levels are weighted equally since we do not focus on a particular decision-making problem, aimed at a specific hierarchical level.
	
	\begin{table}
		\centering
		\caption{Forecasting performance reported for various HF methods in the ``Tourism'' dataset after applying the rolling-origin evaluation approach (average of 61 evaluations of 12 month ahead forecasts). The performance, as measured by MASE, RMSSE, and AMSE, is estimated both per hierarchical level and across all levels.}
		\begin{tabular}{l c c c c c}
			\textbf{Method} & \textbf{Level 0} & \textbf{Level 1} & \textbf{Level 2} & \textbf{Level 3} & \textbf{Average} \\
			\hline
			\multicolumn{6}{c}{MASE} \\
			\hline
			BU              & 1.184            & 1.050            & 0.923            & 0.857            & 1.003 \\
			TD              & 1.048            & 1.297            & 1.124            & 0.978            & 1.112 \\
			COM-SS          & 1.094            & 0.968            & 0.887            & 0.843            & 0.948 \\ COM-SHR & 1.047 & \textbf{0.956} & 0.872 & 0.824 & 0.925 \\
			ML-RF           & 1.045            & 0.964            & 0.859            & 0.812            & 0.920 \\
			ML-XGB          & \textbf{1.043}   & 0.965            & \textbf{0.859}   & \textbf{0.812}   & \textbf{0.920} \\
			\hline
			\multicolumn{6}{c}{RMSSE} \\
			\hline
			BU              & 1.439            & 1.314            & 1.186            & 1.124            & 1.266 \\
			TD              & \textbf{1.238}   & 1.630            & 1.460            & 1.297            & 1.406 \\
			COM-SS          & 1.308            & 1.225            & 1.137            & 1.109            & 1.195 \\
			COM-SHR         & 1.265            & 1.214            & 1.120            & 1.086            & 1.171 \\
			ML-RF           & 1.261            & \textbf{1.208}   & 1.104            & 1.066            & 1.159 \\
			ML-XGB          & 1.255            & 1.208            & \textbf{1.101}   & \textbf{1.064}   & \textbf{1.157} \\
			\hline
			\multicolumn{6}{c}{AMSE} \\
			\hline
			BU              & 1.066            & 0.639            & 0.443            & 0.350            & 0.624 \\
			TD              & 0.845            & 0.594            & 0.404            & 0.341            & 0.546 \\
			COM-SS          & 0.988            & 0.611            & 0.426            & 0.349            & 0.593 \\
			COM-SHR         & 0.935            & 0.599            & 0.417            & 0.337            & 0.572 \\
			ML-RF           & 0.780            & \textbf{0.526}   & 0.366            & 0.319            & 0.498 \\
			ML-XGB          & \textbf{0.779}   & 0.526            & \textbf{0.365}   & \textbf{0.317}   & \textbf{0.497}
		\end{tabular}\label{table:results_t}
	\end{table}

	\begin{table}
		\centering
		\caption{Forecasting performance reported for various HF methods in the ``Sales'' dataset after applying the rolling-origin evaluation approach (average of 330 evaluations of 8 week ahead forecasts). The performance, as measured by MASE, RMSSE, and AMSE, is estimated both per hierarchical level and across all levels.}
		\begin{tabular}{l c c c c}
			\textbf{Method} & \textbf{Level 0} & \textbf{Level 1} & \textbf{Level 2} & \textbf{Average} \\
			\hline
			\multicolumn{5}{c}{MASE} \\
			\hline
			BU              & 0.491            & 0.516            & 0.540            & 0.516 \\
			TD              & 0.522            & 0.785            & 0.971            & 0.759 \\
			COM-SS          & 0.497            & 0.529            & 0.629            & 0.552 \\
			COM-SHR         & 0.495            & 0.520            & 0.542            & 0.519 \\
			ML-RF           & \textbf{0.433}   & 0.449            & \textbf{0.465}   & \textbf{0.449} \\
			ML-XGB          & 0.447            & \textbf{0.447}   & 0.473            & 0.455 \\
			\hline
			\multicolumn{5}{c}{RMSSE} \\
			\hline
			BU              & 0.653            & 0.710            & 0.741            & 0.701 \\
			TD              & 0.684            & 1.118            & 1.314            & 1.039 \\
			COM-SS          & 0.655            & 0.720            & 0.844            & 0.739 \\
			COM-SHR         & 0.654            & 0.713            & 0.742            & 0.703 \\
			ML-RF           & \textbf{0.625}   & \textbf{0.675}   & \textbf{0.703}   & \textbf{0.668} \\
			ML-XGB          & 0.654            & 0.719            & 0.759            & 0.711 \\
			\hline
			\multicolumn{5}{c}{AMSE} \\
			\hline
			BU              & 0.300            & 0.323            & 0.330            & 0.318 \\
			TD              & 0.320            & 0.423            & 0.627            & 0.456 \\
			COM-SS          & 0.301            & 0.327            & 0.372            & 0.334 \\
			COM-SHR         & 0.305            & 0.327            & 0.331            & 0.321 \\
			ML-RF           & \textbf{0.290}   & 0.312            & 0.308            & 0.303 \\
			ML-XGB          & 0.308            & \textbf{0.301}   & \textbf{0.299}   & \textbf{0.303}
		\end{tabular}\label{table:results_s}
	\end{table}
	
	Before proceeding with the evaluation of the results, we highlight that two of the benchmarks considered, namely COM-SS and COM-SHR, are considered state-of-the-art in the field of hierarchical time series forecasting as they have been proven to significantly improve the base forecasts provided to them as input. Moreover, although much more simplistic in nature, the BU and TD methods are highly competitive and, in some applications, difficult benchmarks to beat. Thus, further improving the performance of HF based on ML approaches becomes a promising, yet challenging task.
	
	The results for the Tourism data presented in Table~\ref{table:results_t} show that, on average, ML-XGB is the most accurate HF method in terms of MASE, doing slightly better than ML-RF\@. Specifically, ML-XGB is 17\% and 8\% more accurate on average when compared to the TD and the BU method, respectively, being also 3\% and 0.6\% more precise than the COM-SS and COM-SHR methods. The same stands in general for the individual hierarchical levels, with the exceptions of the TD method for which results are comparable to the ones of the ML methods at the top level, as well as the COM-SHR that displays the best performance at Level 1. This can be partially explained by reviewing the particularities of these two methods: TD builds on the base forecasts produced for the top level of the hierarchy, thus omitting any information provided from the rest of the series, while COM-SHR combines the forecasts from all the series of the hierarchy in a linear way. As a result, if the fully aggregated series is predictable enough, the TD method will provide accurate results at Level 0. Accordingly, if the information required for accurately predicting a level in the middle of the hierarchy, like Level 1, is not complicated and sufficiently provided by the neighboring levels (Levels 0 and 2), COM-SS and COM-SHR will result in improved forecasting accuracy. Note however that COM-SS is always outperformed by COM-SHR due to the latter incorporating information about the correlation structure of the series.
	
	The results are similar in terms of RMSSE, with just two differences worth reporting. First, in this case, the performance of the TD method at the top level is not only comparable to the one of the ML methods, but actually better by about 2\%. However, TD continues to produce significantly less accurate results for the rest of the hierarchical levels. Second, at Level 1, COM-SHR is no longer the best performing method, being outperformed by both ML approach to a similar extent. Thus, we conclude that ML approaches are generally better in approximating the mean of the future values of the series than their median, a phenomenon which can be possibly attributed to the way these methods learn: Both RF and XGB are optimized by minimizing the sum of squared errors produced. Thus, these models learn how to properly approximate the mean and not necessarily the median of the series being predicted.
	
	This last argument may also explain the bias reported for each method, as measured by AMSE\@. Given that mean squared error can be decomposed into a bias and an accuracy term \cite{Taieb7064712}, both ML-RF and ML-XGB are indirectly trained so that they minimize the bias of the reconciled forecasts. In this regard, in contrast to MASE and RMSSE, the ML HF methods always provide significantly less biased forecasts than the benchmarks, especially for the higher levels of the hierarchy. In particular, ML-XGB, the best performing method in terms of AMSE, is on average 15\% better than the benchmark methods, being also less biased by 8\% for the bottom level, 18\% for the top level, and 14\% for the two levels in the middle. Observe also that the worst performing method in terms of AMSE is the BU, with the TD doing also much better than the COM-SS and COM-SHR methods. This indicates that when the base forecasts produced at the bottom level of the hierarchy are biased, reconciliation methods should put more emphasis on the top level where forecasts are more likely to be robust and, therefore, less biased. On the other hand, the fact that ML methods, which exploit the base forecasts produced at all hierarchical levels in a similar fashion to COM-SS and COM-SHR, are still able to provide unbiased results, highlights the potential of dynamic, non-linear reconciliation approaches.
	
	The results are even more encouraging for the case of the Sales dataset. According to MASE, the ML-RF method is considered the most accurate approach on average, being also the best HF method for all levels apart from Level 1. However, even at Level 1, ML-RF is outperformed only to a small extent by ML-XGB, which is also a ML approach. Moreover, in this dataset, the differences reported between the ML methods and the benchmarks are always significant, with the improvements being around 14\% at the top level, 21\% at the middle, and 26\% at the bottom. In other words, not only the improvements reported for the Sales dataset are greater than those of the Tourism dataset, but can be also observed across all levels, becoming more significant for the lower levels of the hierarchy. This could be due to the major differences reported in the Sales dataset between the retailers, meaning that combining the base forecasts from the complete hierarchy to produce forecasts for a particular series is inappropriate when the series do not share the same patterns, at least to some extent. On the contrary, the results highlight that when a ML HF method is utilized for this purpose, being able to selectively combine the base forecasts, the information from the complete hierarchy could still be relevant. This conclusion is also supported by observing that COM-SS and COM-SHR do similarly in terms of MASE to the relatively much simpler BU method.
	
	The results of MASE are in a general agreement to those reported for the case of the RMSSE. Again, ML-RF, the best performing ML HF method, outperforms all of the benchmarks, with the improvements reported being higher for the lower levels of the hierarchy (6\%, 10\%, and 20\% on average for Levels 0,1 and 2, respectively). However, ML-XGB manages to provide slightly less biased results than ML-RF at all levels apart from the top one. Again, the differences between the two ML approaches are small, with their performance being also much better than that of the benchmarks. For example, according to AMSE, ML-RF is on average 6\% less biased than the benchmarks at the top level, 14\% at the middle level, and 18\% at the bottom level.
	
	\begin{figure}
		\centering
		\caption{The accuracy of methods on sales dataset at different levels}
		\includegraphics[width=\textwidth]{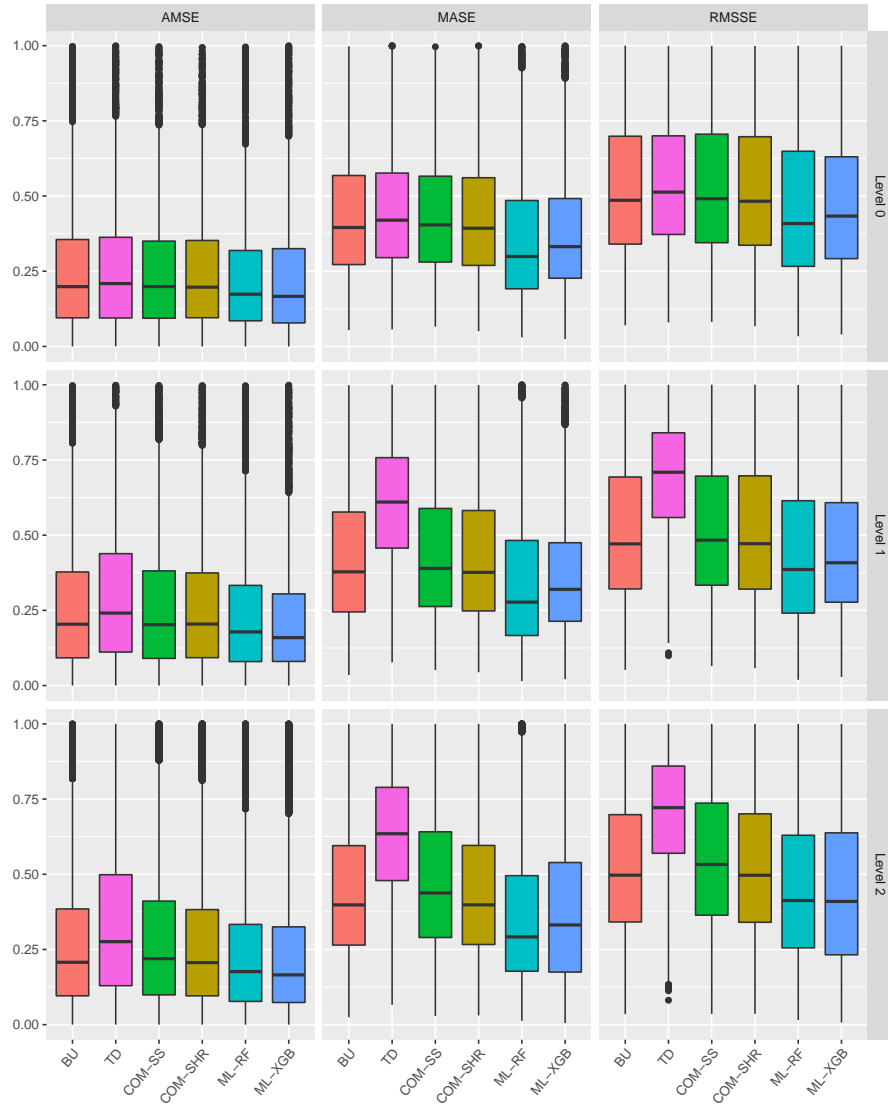}
		\label{ACCAll}
	\end{figure}
	
	Figure~\ref{ACCAll} provides further insight about the relative accuracy of the HF methods for 55 hierarchical sales data at different levels in terms of MASE, AMSE, and RMSSE\@. It demonstrates that both ML HF methods generate more accurate forecasts than their counterparts with ML-RF being the top performing method in terms of MASE\@. While ML-XGB has performed more consistently across different series at Levels 0 and 1, the ML-RF method has generated more consistent forecasts at Level 2. This notion also holds for RMSSE\@. This might be due to different features of time series, such as seasonality, entropy, and the trend of the base time series \cite{abolghasemi2019machine,abolghasemi2020demand}. Finally, it is apparent that the RF and XGB methods performed quite similarly in terms of AMSE.
	
	By summarizing the results of both datasets, the following conclusions can be drawn:
	
	\begin{itemize}
		\item ML HF methods, combining the base forecasts for the complete hierarchy in a non-linear way, provide on average significantly better forecasts than existing methods, both in terms of accuracy and bias. Whether these results can be generalized to other data sets remains to be seen.
		
		\item The information of other series at different levels of the hierarchy (cross-sectional information) can be useful in forecasting the future values of a series regardless of the reconciliation methodology used.
		
		\item The expected improvements from using a ML HF method instead of the existing linear methods are higher when the series in the hierarchy are characterized by different patterns. The greater the differences between the series, at all levels, the higher the potential of using a selective, non-linear reconciliation approach.
		
		\item When a particular ML framework is considered for reconciling the base forecasts produced for an hierarchy, the algorithm selected for determining the combination weights of these forecasts does not greatly affect the final results. Note however that this conclusion is drawn based on an experiment where two decision tree algorithms are used, both utilizing the same framework for performing the reconciliation. Thus, further investigation is required to confirm that this is also the case (i) when different types of ML algorithms (e.g., NNs and SVMs) are used for combining the base forecasts and (ii) different reconciliation frameworks are utilized.
	\end{itemize}
	
	\section{Conclusions}\label{hierarchicalconclusion}
	
	The challenge of hierarchical forecast reconciliation, to produce coherent forecasts across the various hierarchical levels, has so far been tackled with various linear approaches. Early solutions focused on producing forecasts at a single aggregation level with the forecasts of the other levels being derived by aggregation/disaggregation, thus essentially avoiding the incoherence issue. Current state-of-the-art solutions linearly combine the forecasts across all levels. In this study, we have proposed the use of non-linear combination approaches to achieve reconciliation using ML methods.
	
	Our results suggest that, on average, the proposed hierarchical reconciliation approaches based on ML perform well in practice, both in terms of forecast accuracy and bias. Not only can they outperform simple hierarchical approaches, such as BU and TD, but they also show improvements over robust state-of-the-art linear combination approaches. The good performance of HF ML is more evident on the Sales dataset compared to the Tourism dataset, possibly due to the importance of the bottom-level information where our algorithm primarily focuses. The promising empirical results are driven from the design of our approach. HF ML not only results in consistent forecasts across aggregation levels, as is the case with more traditional hierarchical approaches, but also explicitly takes into account the out-of-sample forecast accuracy. The derived combination weights of the HF ML approach provide a selective pooling of the forecasts across the various aggregation levels.
	
	It would be interesting to explore if our insights stand for other ML methods and other data sets. In the following, we discuss additional, alternative paths for future investigation.
	
	\begin{itemize}
		\item In this study, we focused on the case of cross-sectional hierarchical structures. However, forecasting with hierarchies has been extended to the temporal and the cross-temporal dimensions \cite{athanasopoulos2017forecasting,KOURENTZES2019393,Spiliotis2020-hj}. Future work could apply our approach to these dimensions as well and benchmark against standard, linear reconciliation approaches. One challenge, though, has to do with the size of the task, especially in the cross-temporal domain, and the ability to apply the ML approaches described here when the time series are not long enough.
		
		\item Our approach focused on optimizing the performance of the bottom-level series, building $m_k$ models in total. Further research could generalize this optimization objective to other (or multiple) levels of aggregation.
		
		\item We showed that HF ML approaches perform better in the case of point forecasting. Future research could extend our results to include evaluation on the forecast uncertainty \cite{Jeon2019-xo}.
		
		\item Our empirical study included two datasets, sampled in monthly and weekly frequencies. We expect that the performance improvements observed by applying non-linear approaches to hierarchical forecast reconciliation would be amplified for higher data frequencies (e.g., daily or hourly).
		
		\item Despite the improved forecasting performance, the computational complexity should be also examined. It is important to trade-off any gains on the forecast accuracy against additional computational cost/resources \cite{Gilliland2019-xl,Nikolopoulos2018-wa}.
	\end{itemize}
	
	\bibliographystyle{splncs04.bst}
	\bibliography{MLHTS.bib}
	
\end{document}